\documentclass[10pt,twocolumn,letterpaper]{article}

\usepackage{iccv}
\usepackage{times}
\usepackage{epsfig}
\usepackage{graphicx}
\usepackage{amsmath}
\usepackage{amssymb}

\usepackage[pagebackref=true,breaklinks=true,letterpaper=true,colorlinks,bookmarks=false]{hyperref}

\iccvfinalcopy 

\def\iccvPaperID{9976} 

\ificcvfinal\fi

\begin{document}

\title{Consumer Image Quality Prediction using Recurrent Neural Networks for Spatial Pooling}

\author{Jari Korhonen, Yicheng Su\\
Shenzhen University\\
Shenzhen, China\\
{\tt\small jari.t.korhonen@ieee.org}
\and
Junyong You\\
Norwegian Research Center\\
Bergen, Norway
}

\maketitle
\ificcvfinal\thispagestyle{empty}\fi

\begin{abstract}
   Promising results for subjective image quality prediction have been achieved during the past few years by using convolutional neural networks (CNN). However, the use of CNNs for high resolution image quality assessment remains a challenge, since typical CNN architectures have been designed for small resolution input images. In this study, we propose an image quality model that attempts to mimic the attention mechanism of human visual system (HVS) by using a recurrent neural network (RNN) for spatial pooling of the features extracted from different spatial areas (patches) by a deep CNN-based feature extractor. The experimental study, conducted by using images with different resolutions from two recently published image quality datasets, indicates that the quality prediction accuracy of the proposed method is competitive against benchmark models representing the state-of-the-art, and the proposed method also performs consistently on different resolution versions of the same dataset.
\end{abstract}

\section{Introduction}

Blind image quality assessment (BIQA) refers to techniques for predicting subjectively perceived image quality directly from the digital input image. There are several interesting and useful applications of BIQA, ranging from automatic feedback provided to users about the quality of their photos to quality labeling in photo sharing platforms. BIQA methods could even be employed as a cost function for image enhancement algorithms aiming to improve image quality without user supervision. During the past few years, BIQA has attracted considerable attention in the research community. 

The conventional BIQA methods use low-level hand-crafted features extracted from the test image and then fed to a regression model to predict the mean opinion score (MOS) ~\cite{cornia,brisque,niqe,friquee,higrade}. Recently, several BIQA methods based on deep convolutional neural networks (CNNs) have also been proposed ~\cite{kang,bosse,deepbiq,deeprn,koniq10k,spaq,paq2piq,wsp}. The recent studies indicate that CNN-based BIQA models are substantially more accurate than the conventional BIQA methods ~\cite{deepbiq,deeprn,koniq10k,spaq,paq2piq,wsp}.

Since deep CNN architectures are typically very complex, comprising even hundreds of layers, the resolution of input image is usually relatively small. For classification, images can be downscaled in most cases without problems, because the crucial elements (such as edge corners and line tips) making the object recognizable can survive downscaling. However, downscaling may remove features that are essential to determine the perceived image quality, such as fine-grain noise or blurriness~\cite{deepbiq,deeprn}. Therefore, image downscaling may be harmful for BIQA.

There are two basic alternatives to use CNN for BIQA on large resolution images. The first option is patch-based processing: input images are divided into patches, CNN is used as a feature extractor to obtain feature vectors for each patch, and then the feature vectors are combined (e.g. via averaging) to form a feature vector representing the full image, used as input to a regressor (e.g. support vector machine or neural network)~\cite{deepbiq,paq2piq}. The second option is to use arbitrarily sized images as input to the convolution network, and then to obtain the feature vector via global pooling (e.g. global average pooling); in this way, the resulting feature vector has fixed length regardless of the input image resolution, and it can be used directly as input to the regression layer~\cite{deeprn,koniq10k,wsp}.

In this paper, we propose to use recurrent neural network (RNN) for spatial pooling of feature vectors extracted from patches using a CNN-based feature extractor. We assume that reordering the patches appropriately can roughly mimic the visual attention process of human visual system (HVS), where some regions (patches) of the image attract more attention than the others. To improve the performance of the technique, we use two versions of the image as input: higher and lower resolution.

We have compared the proposed technique against state-of-the-art CNN-based BIQA models. Our results show that in many scenarios, the proposed method outperforms the best prior models, when similar test configuration is used for training and testing the proposed model and the benchmark models. We have also conducted an ablation study to investigate the impact of using RNN and multi-scale input in the proposed model.

The rest of the paper is organized as follows. In Section 2, we summarize the relevant related work. In Section 3, we explain the proposed model in detail. In Section 4, we describe the experimental study, demonstrating the performance of the proposed techniques in comparison with the most prominent methods known from the prior art. Finally, the concluding remarks are given in Section 5.

\section{Related research}

Due to the wide range of different possible visual quality distortions, such as sensor noise, motion blur, over- and underexposure, as well as their interplay, development of accurate models for BIQA is a very challenging research problem. The first successful BIQA models use learned or hand-crafted features, often based on natural scene statistics (NSS), and a learning-based regression model to predict MOS~\cite{cornia,brisque,niqe,friquee,higrade}. Support vector regression (SVR) is the most common regression method for conventional BIQA, but models with other regressors, such as random forest and neural networks, have also been studied.

During the past few years, several CNN-based BIQA models have been proposed. Unfortunately, deep CNNs often require a large amount of data (even millions of images) when trained from scratch, and until to date, the publicly available image quality datasets annotated with ground truth subjective MOS have not been sufficiently large to train deep CNNs. This problem has been circumvented by using CNN architectures pre-trained for image classification using large classification datasets, such as ImageNet~\cite{imagenet}, re-purposed for BIQA via transfer learning.

Since image downscaling alters or removes some of the characteristics that are essential for image quality (e.g. high frequency noise), CNN-based BIQA models cannot rely on downscaling to process images larger than the input size of the used CNN model ~\cite{deepbiq,deeprn}. The most obvious option is to divide input image into patches and use each patch as input to CNN separately. Then, MOS for the full image can be predicted by either averaging the MOS for each patch representing the same image, or applying average feature pooling to obtain averaged feature vector as input to SVR to predict MOS~\cite{deepbiq}.

An alternative approach is to apply global pooling to feature maps for dimensionality reduction. Following this approach, images of different resolutions can be used as input, because convolution operation works on images of arbitrary resolution; only the fully connected (FC) layer requires a fixed length feature vector. The simplest methods for global pooling are global average pooling (GAP) and global median pooling (GMP). GAP is used e.g. in the recently proposed KonCept512 model ~\cite{koniq10k}. In~\cite{deeprn}, spatial pyramid pooling (SPP), known from image classification, was employed in a CNN-based BIQA model. However, this method may be cumbersome for datasets containing images of different resolutions, since many deep learning frameworks do not support input images with arbitrary resolution.

Due to the attention mechanism of HVS, some regions of an image typically attract more attention than the others. Therefore, it is expected that an attention mechanism could be useful for weighting different regions of the image according to their relative perceptual importance. In~\cite{saliencyiqa_mtap,saliencyiqa_eusipco}, a saliency model is integrated in patch based BIQA models: only the patches representing salient regions are considered, when the predicted MOS is computed. PaQ-2-PiQ model~\cite{paq2piq} uses region of interest (RoI) pooling to emphasize regions of interest. The weakness of this approach is that it requires training images with MOS for both full image and RoI. In~\cite{wsp}, an attention unit is used between the baseline convolutional network and the FC layer for regression. The attention unit uses 1x1 convolution to compute weight maps for each feature map, and fast normalized fusion operation combines the weight and feature maps. 

RNNs have been widely used for solving classification and sequence-to-sequence conversion problems when the input is a sequence of feature vectors. RNNs have been popular in various applications, especially in speech recognition and natural language processing (NLP). Typical RNNs are built by using either long-short term memory (LSTM) units~\cite{lstm} or gated recurrent units ~\cite{gru} as basic building blocks. RNNs have been used for temporal processing in video quality models ~\cite{3dcnnvqa,vsfa}, where feature vectors obtained from video frames naturally form a time sequence. However, we are not aware of image quality models using RNNs, although we may assume that the gaze pattern also forms a time sequence as a human observer looks at an image. This has motivated us to apply RNN for spatial pooling and regression in a patch based BIQA model.

Since virtually all the best performing BIQA models are learning based models, the used training dataset is crucial for the model performance. Several image quality databases have been published during the past two decades, covering both artificial and natural distortions. In this study, our focus is on user generated content (UGC) with authentic distortions. The best-known natural image quality datasets include LIVE in the Wild Image Quality Challenge (CLIVE)~\cite{clive}, and KonIQ-10k~\cite{koniq10k}.

CLIVE is the first large scale image quality dataset focused on natural distortions, published in 2015~\cite{clive}. Relatively small resolution is used, as most of the images in CLIVE have resolution of $500\times500$ pixels. The dataset contains 1,162 images, each image rated in average by 175 test users in an online crowdsourcing study. KonIQ-1k, published in 2019, is a substantially larger dataset of 10,073 images, each image rated in an online study by more than 100 test users in average. The image resolution in KonIQ-10k is $1,024\times768$ pixels.

Most recently, SPAQ~\cite{spaq} and LIVE-FB~\cite{paq2piq} datasets have been published in 2020. SPAQ contains 11,125 smartphone photos in different resolutions (mostly large resolution), taken with smartphones. Unlike in the case of CLIVE and KonIQ-10k, the user ratings for SPAQ were collected in a laboratory environment. In addition to quality scores, additional information was also collected, such as categorical labels for the images. LIVE-FB is even larger dataset with nearly 40,000 images and more than 100,000 patches extracted from the images. LIVE-FB includes images with different resolutions, but most of the images are relatively small.

Even the largest image datasets with authentic distortions (LIVE-FB, KonIQ-10k, and SPAQ) are small in comparison with the largest image classification datasets. Nevertheless, they have been used for training BIQA models via transfer learning with success. On KonIQ-10k, the state-of-the-art results have been achieved by KonCept512~\cite{koniq10k} and WSP~\cite{wsp}, using InceptionResNetV2 and ResNet-101 as baseline CNN models, respectively. On SPAQ, the authors have achieved good results by using ResNet-50 as baseline model, fine-tuned into a patch-based BIQA model. Further improvement in the results have been reported by using auxiliary EXIF data as additional input to the image quality predictor~\cite{spaq}.

Even though several different architectures have been proposed for predicting subjective quality of images with both artificial and natural distortions, those methods have been mostly trained and tested on images with relatively small resolutions. There is some evidence that CNN-based quality models with global pooling may not perform optimally on large resolution images: developers of KonCept512 have observed that their model achieved better results on KonIQ-10k dataset, when the original images were downscaled to $512\times384$ pixels from the original resolution of $1024\times768$ pixels~\cite{koniq10k}. Therefore, it seems that there is room of improvement in BIQA models, especially when large resolution images are concerned.

\begin{figure*}
\begin{center}
\includegraphics[width=0.8\linewidth]{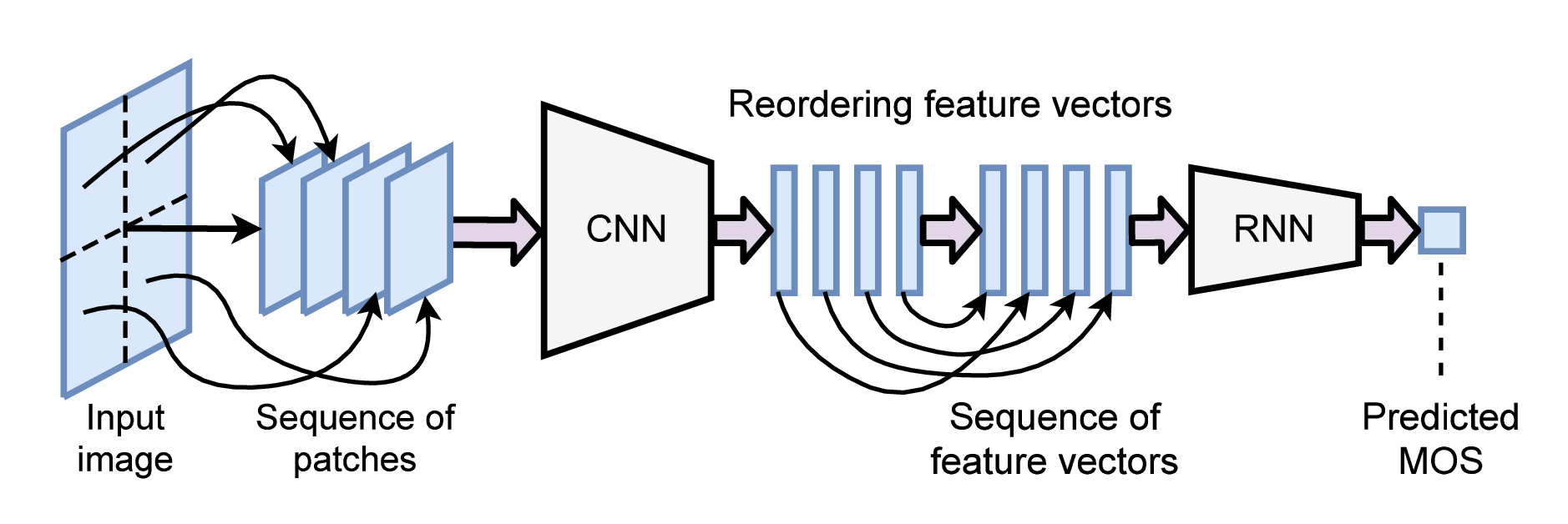}
\end{center}
   \caption{Conceptual illustration of the proposed model.}
\label{fig:fullmodel}
\end{figure*}

\section{Proposed model}

In this study, we propose two main contributions to improve the state-of-the-art in BIQA for high resolution images. First, we use RNN for spatial pooling and regression to predict MOS scores from a sequence of feature vectors, obtained from patches by using a CNN as a feature extractor. Second, we use multiresolution processing by extracting patches from the large resolution image, as well as from the downscaled version of the same image.

\subsection{Using RNN for spatial pooling and regression}

The basic principle of proposed model is illustrated in Figure~\ref{fig:fullmodel}. We first divide the image in patches of the same size as the input image accepted by the CNN model (e.g., when ResNet-50 architecture is used as backbone, the size is $224\times224$ pixels). The patches are sampled with regular intervals to cover the whole image; therefore, the patches will partially overlap, unless the height and width of the image are both divisible by 224. Assuming that the input image resolution $W\times H$ is larger than $224\times224$ pixels, horizontal stride $s_h$ and vertical stride $s_v$ can be computed using Equation (1):

\begin{equation}
s_{h}={\lfloor\frac{W-224}{\lceil{W/224}\rceil-1}\rfloor},
s_{v}={\lfloor\frac{H-224}{\lceil{H/224}\rceil-1}\rfloor}
\end{equation}

The convolution layers of the CNN model are used to extract the feature vectors for every patch. The obtained sequence of feature vectors is then used as input to the RNN regression model to predict MOS. To get better results, the CNN model can be fine-tuned for BIQA task via transfer learning by using small resolution input images (patches) for training.

When time sequences are concerned, the order of feature vectors usually follows their time order. In our model, we aim to mimic the gaze pattern of a human observer. Therefore, we try to predict the relative importance of the patches and reorder the feature vectors accordingly, before feeding them to the RNN unit. There are several different alternative criteria that can be used to decide the order, e.g. spatial activity index of the patches. In our model, we compute the spatial activity index (SI) as defined in~\cite{si} for all the patches, and then reordered the feature vectors in ascending order, according to their SI. We assume that a patch containing interesting details produces higher SI than a patch that lacks of details. Therefore, we may consider SI as a rough estimator of relative salience of the patches. The definition of SI is given in Equation (2), where $\sigma$ denotes standard deviation, $Sobel$ denotes Sobel filter operation, and $Y$ is monochrome version of the input patch:

\begin{equation}
SI={\sigma(Sobel(Y))}
\end{equation}

Different alternative architectures for the RNN unit can be used. Most RNN architectures use long-short term memory (LSTM) or gated recurrent units (GRU) as basic building blocks. From prior studies, it is known that GRU often achieves similar accuracy to LSTM, but with lower complexity, since GRU uses fewer learnable parameters. Therefore, we have chosen to use GRU in our RNN model. It has also been suggested that chaining recurrent layers can improve the model performance. Hence, we used four GRU layers in our model, with the number of hidden units as 256, 128, 64, and 32, respectively. 

\begin{figure}[t]
	\begin{center}
		\includegraphics[width=1\linewidth]{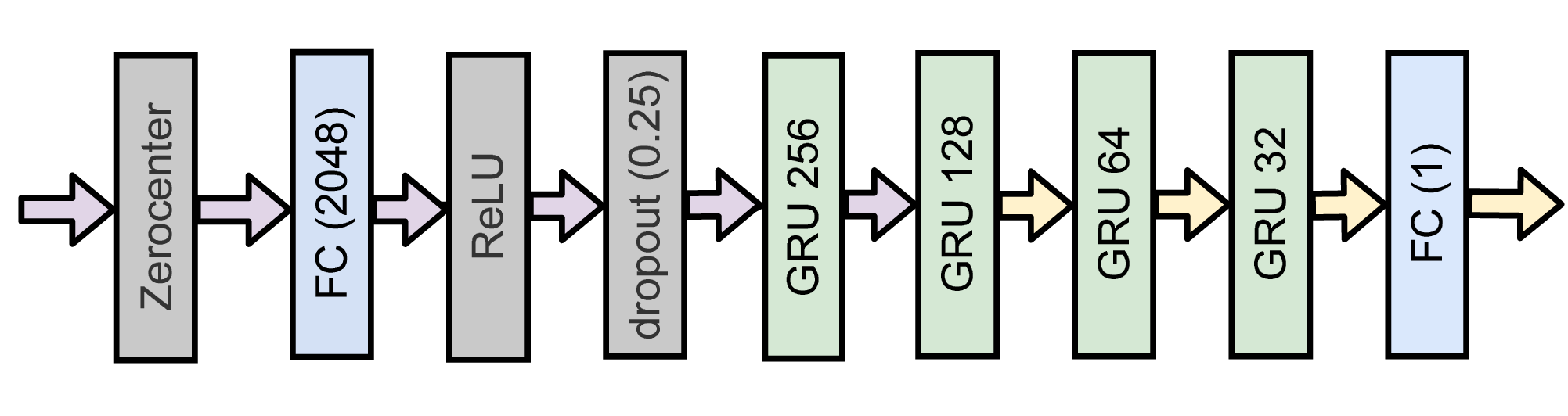}
	\end{center}
	\caption{Proposed RNN model.}	
	~\label{fig:rnnmodel}
\end{figure}

In our model, only the first two GRU layers process sequential data. Single feature vectors are used as inputs to the third and the fourth GRU layers with 64 and 32 hidden units, respectively, so they are only used for reducing the length of the feature vector. We observed that this approach stabilized the training process and produces better results than using GRU layers only for processing sequential data. We also used zerocenter normalization on the input data and FC feed-forward layer (with the same number of outputs as their inputs), combined with related ReLU and dropout layers (dropout rate 0.25) in front of the GRU layers. The last FC layer has one output, namely the predicted MOS. The architecture of the proposed RNN unit is illustrated in Figure~\ref{fig:rnnmodel}. Purple arrows denote flow of sequential data, and yellow arrows denote flow of single feature vectors.

\subsection{Multiresolution processing}

\begin{figure}[t]
	\begin{center}
		\includegraphics[width=0.8\linewidth]{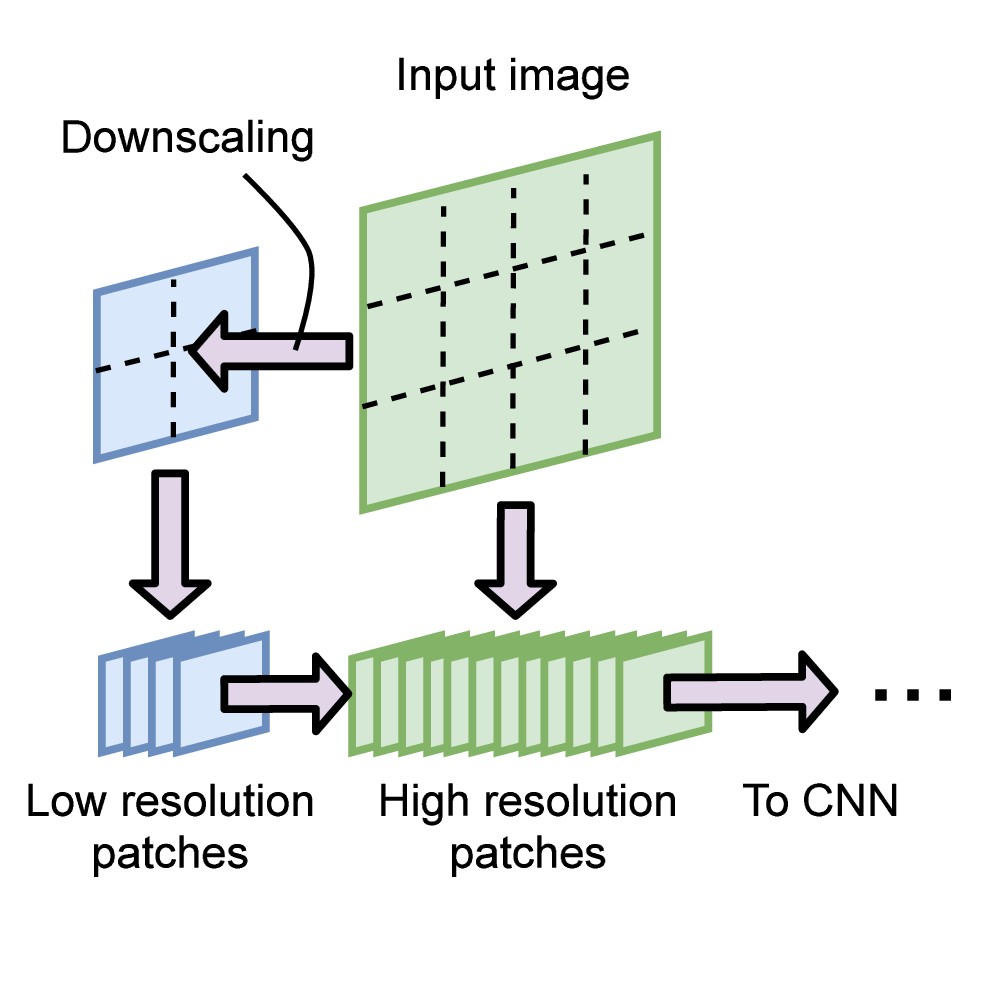}
	\end{center}
	\caption{Multiresolution processing of the input image.}	
	~\label{fig:multireso}
\end{figure}

Different types of quality distortions can appear at different levels of the image. High level distortions, such as under- or overexposure, as well as motion blur, can be observed best in the downscaled version of the image. On the other hand, low level distortions, such as sensor noise or minor out-of-focus blur, are only apparent in the high resolution version. We expect that HVS first focus at higher level characteristics of the image, and then at the lower level details. This is our motivation for multiresolution processing: we first downscale the image and feed the patches from the low resolution image to the CNN, followed by the patches from the high resolution version of the image. The feature vectors obtained for the patches from the lower and higher resolution input image are rearranged separately, as explained in the previous subsection. The basic principle of the method is illustrated in Figure~\ref{fig:multireso}.

If the resolution of the input image is larger than the native resolution of the display device, it is reasonable first to downscale the input image to the size of the displayed image. Otherwise, the predicted MOS may be influenced by fine grain degradations that are invisible to the end users. In our experiments, we have tried different resolution versions of one of the test datasets (SPAQ). The details of the test procedure are given in Section 4.

It should be noted that even though multi-scale models for image classification and quality prediction have been proposed in the prior art~\cite{multiscale,mlsp}, those models typically exploit intermediate \textit{feature maps} of different scales, not re-scaled input patches. We are not aware of prior BIQA methods using different resolutions at the input image level.

\section{Experiments}

In the experimental part of the study, we have used CLIVE dataset with small resolution images ($500\times500$ pixels) for fine-tuning the feature extractor (CNN model). Datasets with higher resolution images, namely KonIQ-10k and SPAQ, have been used for testing the full model. We have conducted an ablation study to investigate the impact of the proposed contributions separately. We have also compared the performance of the full proposed model against state-of-the-art models, using both KonIQ-10k and SPAQ datasets for training and testing separately, as well as in a cross-dataset scenario, involving both datasets.

\subsection{Used datasets}

We did not consider traditional image quality datasets with artificial distortions (e.g. CQIS~\cite{cqis} and TID2013~\cite{tid2013}) for this study, since our focus is on UGC photos with natural distortions. We identified four potentially suitable large scale natural image quality datasets: CLIVE~\cite{clive,clivedataset}, KoNIQ-10k~\cite{koniq10k}, SPAQ~\cite{spaq}, and LIVE-FB~\cite{paq2piq}. Among these datasets, we decided to exclude FLIVE, since its distribution of different resolutions was considered unsuitable for our study.

KonIQ-10k and SPAQ are relatively large datasets with over 10,000 images, nearly ten times the number of images in CLIVE. The resolution of images in CLIVE is smaller than in the two other datasets. Based on the different characteristics of the datasets, we assume that CLIVE would be appropriate for training the CNN model for feature extraction, whereas the two other datasets would be more appropriate for training and testing the RNN model for high resolution images.

KonIQ-10k and SPAQ datasets have roughly similar number of images. Although there are some differences in subjective testing methodologies, we consider those datasets sufficiently similar for cross-dataset experiments. The MOS histograms show that the subjective quality distribution of KonIQ-10k is somewhat biased towards high quality images, whereas SPAQ has more even distribution of different quality images. However, both datasets have been used successfully in the prior art for training and testing image quality models, and therefore we assume that those differences do not have a crucial importance.

\subsection{Training the {CNN} feature extractor}

We have chosen ResNet-50 as the baseline architecture for our feature extractor. Since the available pretrained ResNet-50 model is trained for image classification with ImageNet dataset, we have fine-tuned the model for BIQA task via transfer learning, following the standard approach known from many related studies. We removed the FC and SoftMax layers used for classification and replaced them by a new head consisting of an FC layer with five outputs, representing the probabilistic distribution of votes for the five quality categories, followed by SoftMax layer and regression layer using cross-entropy as loss function. The weights for the first 36 layers, covering the first two convolution blocks of ResNet-50, were frozen. After re-training, the head was removed, and the output from the average pooling layer is used as a feature vector, containing 2,048 elements.

We used CLIVE dataset as training dataset for fine-tuning the feature extractor. All except two images in CLIVE dataset have resolution of $500\times500$ pixels. Therefore, we can expect that the subjective quality of $224\times224$ pixel patches extracted from such small test images represent the overall MOS more accurately than patches extracted from high resolution images, where different regions may have very different subjective quality. For example, the background in photos with strong bokeh effect may look blurry, and in this case the patches extracted from the background would be rated with low quality scores, even though a sharp object in the foreground is more definitive for the overall subjective quality. In small resolution images, we expect that such discrepancy of subjective quality between different regions of the image is less apparent.

For training the CNN model, we have followed similar procedure as described in~\cite{cnntlvqm}. We divided each image into partially overlapping patches so that the whole image area is covered. Nine patches from each $500\times500$ pixel image were extracted. Since CLIVE contains 1,162 images, we obtained more than 10,000 patches. We increased the number of patches further by rotating each patch by 90, 180, and 270, resulting in over 40,000 patches in total. In the related work, it has been suggested that BIQA models are more accurate if they are trained to predict the probabilistic distribution of quality categories, instead of predicting MOS directly~\cite{probqualbiqa,deeprn}. Therefore, each patch was assigned a probabilistic representation of the MOS score, consisting of five bins representing different quality categories, computed as truncated Gaussian distribution from the ground truth MOS and its standard deviation. 

Matlab with deep learning toolbox was employed for training the model. Since the standard layers in Matlab do not support cross-entropy for regression, we implemented custom layers for SoftMax and cross-entropy regression. We used stochastic gradient descent with momentum (SGDM) optimizer for training, learning rate set to $0.5\cdot10^{-4}$. To reduce the effect of overfitting, we used batch size of 32, L2 regularization parameter 0.01, and only two epochs. Random shuffling was applied to the training images after every epoch. For the other learning parameters, we used the default settings.

\subsection{Training and testing the RNN model}

Since image processing and inference of deep learning models is relatively slow in Matlab, we converted the CNN feature extractor model into open neural network exchange (ONNX) format, and implemented feature extractor for high resolution images in C++ (Visual Studio), employing OpenCV and ONNX Runtime libraries. The feature extractor divides input images into patches and uses the CNN feature extractor to generate a sequence of feature vectors, as described in Section 3. OpenCV's resampling using pixel area relation method is used for image resizing in multiresolution processing, as it is the recommended method for downscaling. The obtained sequence of feature vectors is saved in a data file. In our experiments, the processing time for one image of resolution $1024\times768$ was about one second, which is acceptable for most of the practical applications.

In KonIQ-10k dataset, all the images have the same resolution ($1024\times768$ pixels). In contrast, SPAQ dataset includes images with different resolutions and aspect ratios, and many images are very large, even over four million pixels. However, for the subjective experiments, the smaller dimension was reduced to 512 pixels, while the aspect ratio was preserved~\cite{spaq}. In our study, we made three versions of the SPAQ dataset by reducing the smaller dimension to 512, 768, and 1,280 pixels. The aspect ratio was preserved: therefore, the larger dimension is not fixed. These datasets are denoted as SPAQ-512, SPAQ-768, and SPAQ-1280, respectively. The largest resolution (SPAQ-1280) is roughly equivalent to a typical computer monitor resolution.

After computing the feature vector sequences for the images in KoNIQ-10k and SPAQ datasets, we used the sequences in Matlab for training and testing the RNN model, described in Section 3, to predict MOS. In the results, we reported MOS prediction accuracy in terms of Spearman rank order correlation coefficient (SCC), Pearson linear correlation coefficient (PCC), and root mean squared error (RMSE), which is the standard practice in the related studies. Note that MOS for both datasets use range 0-100 (raw MOS for SPAQ and MOS Z-scores for KonIQ-10k), but for training the RNN model, we rescaled MOS linearly to range [0,1]. However, the reported RMSE results use the original scale.

For training the RNN model, we used Adam optimizer, as it is commonly known to perform better than SGDM for training RNNs. The initial learning rate was set to $2\times10^{-4}$, reduced by a factor of 0.5 after every epoch. The batch size was set to 16, L2 regularization to $1\times10^{-5}$, gradient decay factor to 0.95, and squared gradient decay factor to 0.9. We trained the model using five epochs. To make training process more resilient to outliers, Huber loss ($\delta=1/9$) was used as loss function, as suggested in~\cite{deeprn}. We used the same training options for all the test scenarios. 

\subsection{Results}

Our first experiment was an ablation study, where we investigated how different proposed mechanisms influence the accuracy of the BIQA model. We assume that the results of cross-dataset experiments reflect the generalizability of the model better than the results on a single dataset. Hence, we conducted the ablation study in the cross-dataset scenario. 

\begin{table}
	\begin{center}
		\begin{tabular}{|l|ccc|}
			\hline
			Test scenario & SCC & PCC & RMSE \\
			\hline\hline
			CNN+Avg+NN & 0.707 & 0.732 & 11.3 \\
			CNN+Mres+Avg+NN & 0.732 & 0.758 & 10.6 \\
			\hline
			CNN+RNN & 0.794 & 0.815 & 10.3 \\
			CNN+Mres+RNN & 0.810 & 0.831 & 9.8 \\
			\hline
		\end{tabular}
	\end{center}
	\caption{Ablation study: training on SPAQ, testing on KoNIQ-10k.}
\end{table}

\begin{table}
	\begin{center}
		\begin{tabular}{|l|ccc|}
			\hline
			Test scenario & SCC & PCC & RMSE \\
			\hline\hline
			CNN+Avg+FC & 0.830 & 0.827 & 12.1 \\
			CNN+Mres+Avg+FC & 0.849 & 0.846 & 11.4 \\
			\hline
			CNN+RNN & 0.845 & 0.841 & 11.6 \\
			CNN+Mres+RNN & 0.860 & 0.855 & 11.0 \\
			\hline
		\end{tabular}
	\end{center}
	\caption{Ablation study: training on KoNIQ-10k, testing on SPAQ.}
\end{table}

As a baseline model, we used simple element-wise averaging to combine the feature vectors for individual patches into a single feature vector, representing the whole image. Then, we used a neural network with zerocenter normalization, FC layer with 256 outputs, ReLU layer, dropout layer with rate 0.25, and finally, FC layer with one output, for predicting MOS. This model is denoted as “CNN+Avg+NN”. Huber loss ($\delta=1/9$) was used as loss function, as also in the proposed RNN-based model. In the next step, we added multiresolution processing, denoted as “CNN+MRes+Avg+NN”, as described in Subsection 3.2. The baseline model is conceptually similar to DeepBIQ with feature average pooling~\cite{deepbiq}, but with different patch sampling strategy, backbone network, and regression model.

The simplest version of the proposed model with RNN is denoted as “CNN+RNN”. This model uses only the patches extracted from the image in the original resolution. 20 patches are extracted from each KonIQ-10k image. For SPAQ-768, the number of patches varies between 16 and 32 (median is 20). The full model, denoted as “CNN+MRes+RNN”, includes also multiresolution processing, where the low resolution image is obtained by downscaling the input image by a factor of 0.5. For KonIQ-10k, there are six patches extracted from the small resolution image, and for SPAQ-768, the number of patches varies between four and eight (median is six).

Tables 1 and 2 show the ablation study results, when SPAQ-768 dataset is used for training and KonIQ-10k for testing, and vice versa. The results show that the proposed method with RNN performs clearly better than the baseline, especially when SPAQ-768 is used for training. Multiresolution processing improves the performance for both baseline and the RNN-based method.

The authors of KonCept512 BIQA model observed that their model works best with images of $512\times384$ pixels. To study the impact of image resizing on our proposed model, we split SPAQ dataset randomly into training and test sets (80\% of the images for training and 20\% for testing), and run the proposed model by using the three versions of the dataset with different resolutions: SPAQ-512, SPAQ-768, and SPAQ-1280.

As shown in Table 3, the results for different resolutions are nearly identical. In fact, the results on larger resolution datasets (SPAQ-768 and SPAQ-1280) are even slightly better than the results on SPAQ-512 (the resolution used in the subjective study). This indicates that the proposed model works well on different resolutions, which is a clear benefit in comparison with methods using global pooling.

\begin{table}
	\begin{center}
		\begin{tabular}{|l|ccc|}
			\hline
			Dataset & SCC & PCC & RMSE \\
			\hline\hline
			SPAQ-512 & 0.917 & 0.925 & 8.05 \\
			SPAQ-768 & 0.920 & 0.927 & 7.95 \\
			SPAQ-1280 & 0.920 & 0.927 & 7.96 \\
			\hline
		\end{tabular}
	\end{center}
	\caption{Test results for the proposed method on different resolution versions of SPAQ. The same random split (80:20) is used for each dataset.}
\end{table}

To compare the proposed model against relevant benchmark methods, we have run eight comparison experiments: two cross-dataset experiments (training with SPAQ and testing with KonIQ-10k, and vice versa), and six experiments with 80:20 random splits into training and test sets on a single dataset (three on SPAQ, and three on KonIQ-10k). We have selected five different BIQA models as benchmarks: BRISQUE~\cite{brisque} and FRIQUEE~\cite{friquee}, representing traditional BIQA models with handcrafted features and SVR as regression model, and DeepBIQ~\cite{deepbiq}, KonCept512~\cite{koniq10k}, and WSP~\cite{wsp}, representing state-of-the-art deep CNN models.

\begin{table}
	\begin{center}
		\begin{tabular}{|l|ccc|}
			\hline
			Model & SCC .& PCC & RMSE \\
			\hline\hline
			BRISQUE & 0.514 & 0.508 & 18.1 \\
			FRIQUEE & 0.646 & 0.670 & 13.2 \\
			\hline
			DeepBIQ & 0.784 & 0.767 & 14.7 \\
			KonCept512 & 0.777 & 0.810 & 9.2 \\
			WSP & 0.782 & 0.815 & 10.4 \\
			Proposed & 0.810 & 0.831 & 9.8 \\
			\hline
		\end{tabular}
	\end{center}
	\caption{Comparison study results when training on SPAQ and testing on KoNIQ-10k.}
\end{table}

\begin{table}
	\begin{center}
		\begin{tabular}{|l|ccc|}
			\hline
			Model & SCC .& PCC & RMSE \\
			\hline\hline
			BRISQUE & 0.677 & 0.658 & 25.0 \\
			FRIQUEE & 0.791 & 0.751 & 14.8 \\
			\hline
			DeepBIQ & 0.653 & 0.665 & 13.2 \\
			KonCept512 & 0.864 & 0.862 & 11.3 \\
			WSP & 0.876 & 0.876 & 11.5 \\
			Proposed & 0.860 & 0.855 & 11.0 \\
			\hline
		\end{tabular}
	\end{center}
	\caption{Comparison study results when training on KoNIQ-10k and testing on SPAQ.}
\end{table}

\begin{table*}
	\begin{center}
		\begin{tabular}{|l|ccc|ccc|ccc|ccc|}
			\hline
			& \multicolumn{3}{|c|}{Split 1} & \multicolumn{3}{|c|}{Split 2} & \multicolumn{3}{|c|}{Split 3} & \multicolumn{3}{|c|}{average} \\
			Model & SCC .& PCC & RMSE & SCC & PCC & RMSE & SCC & PCC & RMSE & SCC & PCC & RMSE \\
			\hline\hline
			BRISQUE & 0.673 & 0.668 & 11.54 & 0.678 & 	0.684 & 11.54 & 0.666 & 0.672 & 11.73 & 0.672 & 0.675 & 11.60 \\
			FRIQUEE & 0.826 & 0.844 & 8.28 & 0.828 & 0.852 & 8.27 & 0.822 & 0.850 & 8.33 & 0.825 & 0.849 & 8.29 \\
			\hline
			DeepBIQ & 0.849 & 0.856 & 7.99 & 0.857 & 0.870 &	7.75 & 0.854 & 0.863 & 7.98 & 0.853 & 0.863 &
			7.91 \\
			KonCept512 & 0.908 & 0.926 & 6.01 & 0.916 & 0.930 & 5.96 & 0.908 & 0.922 & 6.34 & 0.911 & 0.926 &	6.10 \\
			WSP & 0.916 & 0.931 & 5.66 & 0.918 & 0.931 & 5.80 & 0.921 & 0.935 & 5.62 & 0.918 & 0.932 & 5.69 \\
			Proposed & 0.916 & 0.930 & 5.68 & 0.923 & 0.935 & 5.61 & 0.922 & 0.935 & 5.66 & 0.920 & 0.933 & 5.65 \\
			\hline
		\end{tabular}
	\end{center}
	\caption{Comparison study results on KoNIQ-10k for three different random splits for training and testing (80:20).}
\end{table*}

\begin{table*}
	\begin{center}
		\begin{tabular}{|l|ccc|ccc|ccc|ccc|}
			\hline
			& \multicolumn{3}{|c|}{Split 1} & \multicolumn{3}{|c|}{Split 2} & \multicolumn{3}{|c|}{Split 3} & \multicolumn{3}{|c|}{average} \\
			Model & SCC .& PCC & RMSE & SCC & PCC & RMSE & SCC & PCC & RMSE & SCC & PCC & RMSE \\
			\hline\hline
			BRISQUE & 0.779 & 0.786 & 12.97 & 0.782 &	0.788 & 12.97 &	0.780 & 0.785 &	12.99 &	0.780
			& 0.786 &12.98 \\
			FRIQUEE & 0.883 & 0.888 & 9.65 & 0.881 &	0.886 &	9.73 & 0.881 & 0.885 & 9.76 & 0.882& 0.886 & 9.71 \\
			\hline
			DeepBIQ & 0.896 & 0.903 & 9.02 & 0.895 & 	0.899 &	9.25 & 0.899 & 0.903 & 9.01 & 0.897 & 0.902 & 9.09 \\
			KonCept512 & 0.910 & 0.917 & 8.65 &	0.910 & 0.914 & 9.01 & 0.915 & 0.917 & 8.81 & 0.912 & 0.916 &	8.82 \\
			WSP & 0.914 & 0.920 & 8.27 & 0.910 & 0.917 & 8.44 & 0.915 & 0.918 & 8.36 & 0.913 & 0.918 &8.36 \\
			Proposed & 0.920 & 0.927 & 7.95 & 0.921 & 	0.927 &	7.96 & 0.922 & 0.926 & 8.05 & 0.921 & 0.927 &	7.99 \\
			\hline
		\end{tabular}
	\end{center}
	\caption{Comparison study results on SPAQ for three different random splits for training and testing (80:20).}
\end{table*}

BRISQUE is a well-established traditional image quality model, implemented e.g. in Matlab image processing toolbox. In our study, we used the built-in Matlab functions for training and testing BRISQUE. In contrast, FRIQUEE is a more recently proposed model based on the “bag of features” approach, and it has achieved state-of-the-art results among the conventional image quality models~\cite{friquee}. Even though FRIQUEE is burdened by its high complexity, we selected it as a benchmark method for our study, since it is the most accurate traditional model. In our study, we used the Matlab implementation provided by the authors for extracting the FRIQUEE features, and SVR implementation in Matlab machine learning toolbox for training and testing the regression model. The original implementation of FRIQUEE crashed when monochrome images were used as input, and therefore we modified the implementation so that the luma-related features are replaced by zeros, if a monochrome image is detected.

We selected the deep benchmark methods representing different approaches for BIQA, including both end-to-end trainable models (KonCept512 and WSP) and patch-based models (DeepBIQ). We did not include models that have not been reported to perform competitively against those three models in related studies. DeepBIQ is a patch-based model using a fine-tuned CNN-based feature extractor and SVR for regression~\cite{deepbiq}. We re-implemented the model in Matlab, following the guidelines of the original publication. However, instead of random sampling, we extracted the patches using the same procedure as we used for the proposed model. In addition, we used feature pooling instead of prediction pooling, since SVR does not scale well for larger datasets, and KoNIQ-10k and SPAQ are substantially larger than CLIVE that was used in the original work. For KonCept512 and WSP, we used the original implementations available from the authors.

Since training of KonCept512 and WSP requires fixed resolution for each mini-batch, we generated $512\times384$ pixels resolution version of both KoNIQ-10k and SPAQ datasets for training and testing KonCept512 and WSP. Portrait images in SPAQ were flipped to preserve aspect ratio as accurately as possible. In contrast, original resolution ($1024\times768$) for KoNIQ-10k and SPAQ-768 version of SPAQ dataset were used for training and testing DeepBIQ and the proposed model. 

The results of the comparison study in cross-dataset scenario are shown in Table 4 for training on SPAQ and testing on KoNIQ-10k, and in Table 5 for training on KoNIQ-10k and testing on SPAQ, respectively. We used similar training protocol as for the ablation study described above. As the results show, the proposed method outperforms the benchmark methods in terms of SCC and PCC, when trained on SPAQ and tested on KoNIQ-10k. In terms of RMSE, KonCept512 performs slightly better. When trained on KoNIQ-10k and tested on SPAQ, the proposed method outperforms the benchmark methods in terms of RMSE; however, KonCept512 and WSP achieve slightly higher correlation (SCC and PCC). It should be noted that KonCept512 and WSP may benefit from the use of fixed resolution.

In the second part of the comparison study, we divided SPAQ and KoNIQ-10k randomly in a training set (80\% of the images) and a testing set (20\% of the images). We repeated the experiment with three different random splits. The results are reported in Table 6 for KoNIQ-10k and Table 7 for SPAQ. In this scenario, the proposed method outperformed all the benchmark methods in average results; however, the difference between the proposed method and WSP is marginal on KoNIQ-10k. The results show that there is no significant deviation between the results when different splits are used: the correlation coefficients for the proposed method, as well as KonCept512 and WSP, fit within a range of less than 0.01. The results for the other methods are also reasonably consistent. Therefore, we assume that our experiment gives a realistic idea of the relative performance of different models. 

In summary, the results implicate that the accuracy of the proposed model is competitive against the relevant models representing state-of-the-art; however, the differences between the best models are relatively small. On the other hand, the main benefit of the proposed method in comparison with end-to-end models is the consistent performance on images with different resolutions. 

\section{Conclusions}

In the recently published studies on BIQA, the best MOS prediction accuracy has been achieved by using end-to-end trainable models. However, those models typically achieve the best accuracy at relatively small resolutions, and their performance on high resolution images is not as convincing. In this paper, we have proposed a patch-based BIQA model, using RNN for spatial pooling and regression. Our results show that the proposed model achieves competitive results on two recently published natural image datasets, and the model works well also on images with relatively high resolution. Therefore, our study supports the hypothesis that patch-based BIQA is a competitive approach for predicting quality of high resolution images. To support reproducibility, implementation of the proposed model is available in \url{http://github.com/jarikorhonen/rnnbiqa}.

\section{Acknowledgements}

This work was supported in part by the National Natural Science Foundation of China under Grant 61772348, and Guangdong "Pearl River Talent Recruitment Program" under Grant JCYJ20200109110410133.

{\small
\bibliographystyle{ieee_fullname}
\bibliography{egbib}

\begin{thebibliography}{10}\itemsep=-1pt

\bibitem{si}
{ITU-T Recommendation P.910}: Subjective video quality assessment methods for
  multimedia applications.
\newblock \url{http://itu.int/rec/T-REC-P.910}, Apr. 2008.

\bibitem{deepbiq}
S. {Bianco}, L. {Celona}, P. {Napoletano}, and R. {Schettini}.
\newblock On the use of deep learning for blind image quality assessment.
\newblock {\em Signal, Image and Video Processing}, 12:355–362, 2018.

\bibitem{bosse}
S. {Bosse}, D. {Maniry}, T. {Wiegand}, and W. {Samek}.
\newblock A deep neural network for image quality assessment.
\newblock In {\em Proc. ICIP'16}, pages 3773--3777, 2016.

\bibitem{saliencyiqa_eusipco}
A. {Chetouani}.
\newblock A blind image quality metric using a selection of relevant patches
  based on convolutional neural network.
\newblock In {\em Proc. EUSIPCO'18}, Rome, Italy, 2018.

\bibitem{gru}
K. {Cho}, B. {van Merrienboer}, C. {G{\"{u}}l{\c{c}}ehre}, D. {Bahdanau}, F.
  {Bougares}, H. {Schwenk}, and Y. {Bengio}.
\newblock Learning phrase representations using {RNN} encoder-decoder for
  statistical machine translation.
\newblock In {\em Proc. {EMNLP'14}}.

\bibitem{spaq}
Y. {Fang}, H. {Zhu}, Y. {Zeng}, K. {Ma}, and Z. {Wang}.
\newblock Perceptual quality assessment of smartphone photography.
\newblock In {\em Proc. CVPR'20}, pages 3677--3686, June 2020.

\bibitem{clivedataset}
D. {Ghadiyaram} and A.~C. {Bovik}.
\newblock {LIVE} in the wild image quality challenge database.
\newblock \url{http://live.ece.utexas.edu/research/ChallengeDB/index.html},
  2015.

\bibitem{clive}
D. {Ghadiyaram} and A.~C. {Bovik}.
\newblock Massive online crowdsourced study of subjective and objective picture
  quality.
\newblock {\em IEEE Transactions on Image Processing}, 25(1):372--387, Jan.
  2016.

\bibitem{friquee}
D. {Ghadiyaram} and A.~C. {Bovik}.
\newblock Perceptual quality prediction on authentically distorted images using
  a bag of features approach.
\newblock {\em Journal of Vision}, 17(1), 2017.

\bibitem{lstm}
S. {Hochreiter} and J. {Schmidhuber}.
\newblock Long short-term memory.
\newblock {\em Neural Computing}, 9(8):1735–1780, Nov. 1997.

\bibitem{mlsp}
V. {Hosu}, B. {Goldl{\"u}cke}, and D. {Saupe}.
\newblock Effective aesthetics prediction with multi-level spatially pooled
  features.
\newblock In {\em Proc. {CVPR'19}}, pages 9375--9383, Long Beach, CA, USA,
  2019.

\bibitem{koniq10k}
V. {Hosu}, H. {Lin}, T. {Sziranyi}, and D. {Saupe}.
\newblock Koniq-10k: An ecologically valid database for deep learning of blind
  image quality assessment.
\newblock {\em IEEE Transactions on Image Processing}, 29(1):4041--4056, Jan.
  2020.

\bibitem{multiscale}
G. {Huang}, D. {Chen}, T. {Li}, F. {Wu}, L. {van der Maaten}, and K.~Q.
  {Weinberger}.
\newblock Multi-scale dense networks for resource efficient image
  classification.
\newblock In {\em Proc. ICLR'18}, Vancouver, BC, Canada, 2018.

\bibitem{saliencyiqa_mtap}
S. {Jia} and Y. {Zhang}.
\newblock Saliency-based deep convolutional neural network for no-reference
  image quality assessment.
\newblock {\em Multimedia Tools and Applications}, 77:14859–14872, 2018.

\bibitem{kang}
L. {Kang}, P. {Ye}, Y. {Li}, and D. {Doermann}.
\newblock Convolutional neural networks for no-reference image quality
  assessment.
\newblock In {\em Proc. CVPR'14}, pages 1733--1740, 2014.

\bibitem{cnntlvqm}
J. {Korhonen}, Y. {Su}, and J. {You}.
\newblock Blind natural video quality prediction via statistical temporal
  features and deep spatial features.
\newblock In {\em Proc. ACM MM'20}, page 3311–3319, 2020.

\bibitem{imagenet}
A. {Krizhevsky}, I {Sutskever}, and G.~E. {Hinton}.
\newblock Imagenet classification with deep convolutional neural networks.
\newblock In {\em Proc. NIPS’12}, Lake Tahoe, NV, USA, 2012.

\bibitem{higrade}
D. {Kundu}, D. {Ghadiyaram}, A.~C. {Bovik}, and B.~L. {Evans}.
\newblock No-reference quality assessment of tone-mapped {HDR} pictures.
\newblock {\em IEEE Transactions on Image Processing}, 26(6):2957--2971, 2017.

\bibitem{cqis}
E.C. {Larson} and D.~M. {Chandler}.
\newblock Most apparent distortion: full-reference image quality assessment and
  the role of strategy.
\newblock {\em Journal of Electronic Imaging}, 19(1), 2010.

\bibitem{vsfa}
D. {Li}, T. {Jiang}, and M. {Jiang}.
\newblock Quality assessment of in-the-wild videos.
\newblock In {\em Proc. ACM MM'19}, page 2351–2359, Nice, France, 2019.

\bibitem{brisque}
A. {Mittal}, A.~K. {Moorthy}, and A.~C. {Bovik}.
\newblock No-reference image quality assessment in the spatial domain.
\newblock {\em IEEE Transactions on Image Processing}, 21(12):4695--4708, 2012.

\bibitem{niqe}
A. {Mittal}, R. {Soundararajan}, and A.~C. {Bovik}.
\newblock Making a “completely blind” image quality analyzer.
\newblock {\em IEEE Signal Processing Letters}, 17(3):209--212, 2013.

\bibitem{tid2013}
N. {Ponomarenko}, L. {Jin}, O. {Ieremeiev}, V. {Lukin}, K. {Egiazarian}, J.
  {Astola}, B. {Vozel}, K. {Chehdi}, M. {Carli}, F. {Battisti}, and C.-C. {Jay
  Kuo}.
\newblock {TID2013:} peculiarities, results and perspectives.
\newblock {\em Signal Processing: Image Communication}, 30(1):57--77, Jan.
  2015.

\bibitem{wsp}
Y. {Su} and J. {Korhonen}.
\newblock Blind natural image quality prediction using convolutional neural
  networks and weighted spatial pooling.
\newblock In {\em Proc. ICIP'20}, pages 191--195, 2020.

\bibitem{deeprn}
D. {Varga}, D. {Saupe}, and T. {Szir{\'a}nyi}.
\newblock {DeepRN}: A content preserving deep architecture for blind image
  quality assessment.
\newblock In {\em Proc. ICME’18}, San Diego, CA, USA, 2018.

\bibitem{cornia}
P. {Ye}, J. {Kumar}, L. {Kang}, and D. {Doermann}.
\newblock Unsupervised feature learning framework for no-reference image
  quality assessment.
\newblock In {\em 2012 IEEE Conference on Computer Vision and Pattern
  Recognition}, pages 1098--1105, 2012.

\bibitem{paq2piq}
Z. {Ying}, H. {Niu}, P. {Gupta}, D. {Mahajan}, D. {Ghadiyaram}, and A.~C.
  {Bovik}.
\newblock From patches to pictures {(PaQ-2-PiQ)}: Mapping the perceptual space
  of picture quality.
\newblock In {\em Proc. CVPR'20}, 2020.

\bibitem{3dcnnvqa}
J. {You} and J. {Korhonen}.
\newblock Deep neural networks for no-reference video quality assessment.
\newblock In {\em Proc. ICIP'19}, Taipei, Taiwan, 2019.

\bibitem{probqualbiqa}
H. {Zeng}, L. {Zhang}, and A.~C. {Bovik}.
\newblock Blind image quality assessment with a probabilistic quality
  representation.
\newblock In {\em Proc. ICIP'18}, pages 609--613, 2018.

\end{thebibliography}
}

\end{document}